\documentclass[10pt,twocolumn,letterpaper]{article}

\usepackage{wacv}              

\usepackage{graphicx}
\usepackage{amsmath}
\usepackage{amssymb}
\usepackage{booktabs}
\newcommand{\methodname}{FreeEdit\xspace}

\usepackage[capitalize]{cleveref}
\crefname{section}{Sec.}{Secs.}
\Crefname{section}{Section}{Sections}
\Crefname{table}{Table}{Tables}
\crefname{table}{Tab.}{Tabs.}

\def\wacvPaperID{1380} 
\def\confName{WACV}
\def\confYear{2025}

\begin{document}

\title{Towards a Training Free Approach for 3D Scene Editing}

\author{\quad Vivek Madhavaram$^{1 
}$  \qquad Shivangana Rawat$^{2}$ \qquad Chaitanya Devaguptapu$^{2}$ \\
\qquad Charu Sharma$^{1}$ 
\qquad Manohar Kaul$^{2}$ \\
{\tt\footnotesize madhavaram.vardhan@research.iiit.ac.in}  \\
$^{1}$ Machine Learning Lab, IIIT Hyderabad, India \enspace\enspace
         $^{2}$ Fujitsu Research India}
\maketitle


\begin{abstract}
   Text driven diffusion models have shown remarkable capabilities in editing images. However, when editing 3D scenes, existing works mostly rely on training a NeRF for 3D editing. Recent NeRF editing methods leverages edit operations by deploying 2D diffusion models and project these edits into 3D space. They require strong positional priors alongside text prompt to identify the edit location. These methods are operational on small 3D scenes and are more generalized to particular scene. They require training for each specific edit and cannot be exploited in real-time edits. To address these limitations, we propose a novel method, \methodname, to make edits in training free manner using mesh representations as a substitute for NeRF. Training-free methods are now a possibility because of the advances in foundation model’s space. We leverage these models to bring a training-free alternative and introduce solutions for insertion, replacement and deletion. We consider insertion, replacement and deletion as basic blocks for performing intricate edits with certain combinations of these operations. Given a text prompt and a 3D scene, our model is capable of identifying what object should be inserted/replaced or deleted and location where edit should be performed. We also introduce a novel algorithm as part of \methodname to find the optimal location on grounding object for placement. We evaluate our model by comparing it with baseline models on a wide range of scenes using quantitative and qualitative metrics and showcase the merits of our method with respect to others. Project page: \url{https://vivekmadhavaram.github.io/FreeEdit_page/}
\end{abstract}

\section{Introduction}
\label{sec:intro}

Existing 3D editing approaches~\cite{instructnerf2023, shahbazi2024inserf,zhuang2023dreameditor} rely heavily on computationally expensive, trained-based methods. Advancements in foundation models have opened up new possibilities for training-free approaches in various 3D understanding tasks~\cite{conceptgraphs,hertz2022prompt,tewel2024trainingfreeconsistenttexttoimagegeneration}. Despite these developments, the task of scene editing has yet to fully leverage such advancements. Existing 3D editing approaches predominantly rely on training-based methods that require training a Neural Radiance Field (NeRF) for each new edit and scene.

The ability to edit large-scale three-dimensional (3D) environments has emerged as a pivotal area recently, driven by applications across Virtual Reality (VR)\cite{8962793, 10.1145/3229147.3229166}, Augmented Reality (AR)~\cite{9288477,5666644}, interior designing ~\cite{8104407,electronics10030245} among others \cite{li2024advances, su15043761}. The ability to dynamically modify and enhance 3D scenes in real time is crucial for creating immersive experiences and personalizing spaces to individual preferences. 


\begin{figure}[t]
  \centering
   \includegraphics[width=\linewidth,trim= 0cm 2cm 0cm 0cm]{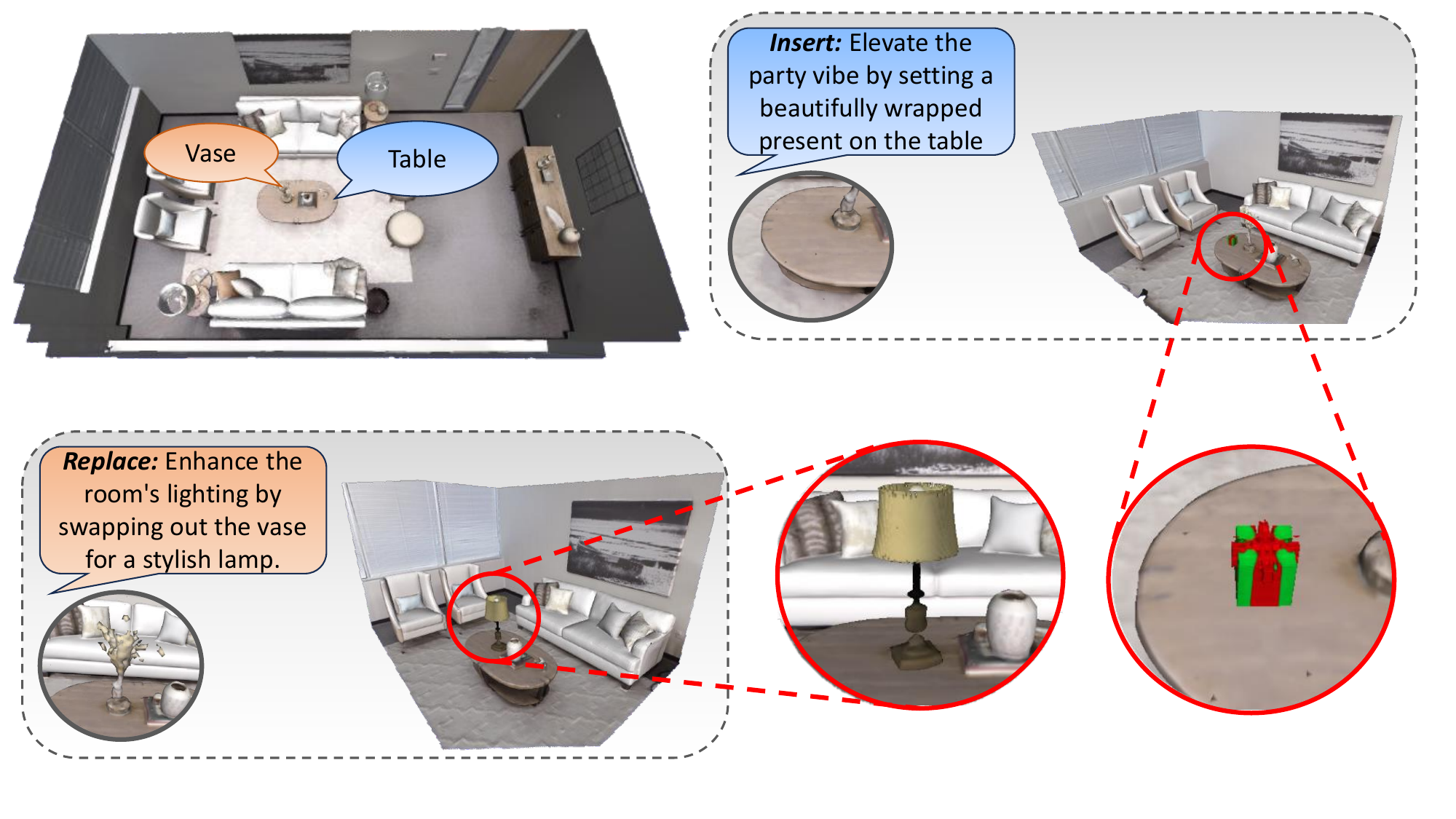}

   \caption{\textbf{Illustration of \methodname for inserting and replacing objects in a complex 3D scene:} Queries in blue and orange illustrate the insertion and replacement prompts provided as input by the user, respectively.}
   \label{fig:teaser}
   \vspace{-0.5cm}
\end{figure}
Most of the existing~\cite{instructnerf2023, shahbazi2024inserf,zhuang2023dreameditor} methods that focus on editing 3D scenes predominantly focus on object-centric modifications or broad scene-level changes such as ambiance adjustments, often neglecting the nuanced requirements of detailed, fine-grained editing. Specifically, the insertion/replacement of objects within complex, multi-object 3D scenes remains inadequately supported. Moreover, these approaches typically require retraining for each new edit, which is both time-consuming and resource-intensive, making them ill-suited for quick or multiple iterative edits. 


To address these challenges, we present \methodname \footnote{The name is inspired by the training-free nature of our approach.}, a \textit{training free} approach for 3D scene editing. Our work primarily focuses on text-guided scene manipulation tasks, including insertion, replacement, and deletion. These fundamental operations serve as building blocks for a wide range of 3D editing tasks. Our main goal is to provide an efficient alternative to expensive training-based methods, leveraging the recent improvements in foundation models to enable near real-time edits within complex scenes comprising numerous objects. 

\methodname operates solely based on text instructions and utilizes mesh representations, enabling efficient editing of room-sized 3D scenes. 
While NeRF-based representations excel in  realistic rendering, our primary focus is on efficiency while maintaining acceptable quality, particularly when editing scenes with multiple objects where NeRF's complexity and need for retraining become bottlenecks. Unlike most NeRF-based editing methods that focus on single-object or small-scale scenes, we target large-scale, room-sized 3D scenes.

Inspired by training-free approaches for various scene understanding tasks~\cite{conceptgraphs,hertz2022prompt,tewel2024trainingfreeconsistenttexttoimagegeneration}, \methodname employs a modular approach for inserting and replacing objects in large-scale 3D scenes (as shown in Figure \ref{fig:teaser}). It supports iterative and real-time edits while eliminating the need for explicit location supervision when inserting objects, instead inferring locations based on user-provided textual instructions. By integrating open-vocabulary segmentation and 3D diffusion models, \methodname broadens the range of recognizable and insertable objects in any given scene, offering a versatile solution for 3D scene editing even when specific 3D assets are unavailable.

\noindent \textbf{Our contributions:} To the best of our knowledge, we are the first to introduce a training-free approach for 3D scene editing. $i)$ We present a framework that enables text-driven object insertion, replacement and deletion in 3D scenes without the prerequisite of training. $ii)$ Our framework is specifically designed to handle large-scale 3D scenes with multiple objects, offering unprecedented flexibility and efficiency in scene customization without the need for positional priors. $iii)$ We introduce an algorithm as part of \methodname to find the optimal location to place the primary object in a scene with minimal intersection. $iv)$ While \methodname is primarily designed for inserting and replacing objects in 3D scenes, by the nature of its design, it also supports interactive object translation, rotation, iterative insertion and object deletion.

It's important to note that while \methodname represents a step towards a more versatile, training-free 3D scene editing, its current implementation focuses on a specific set of object-level manipulations. We acknowledge that this approach may not encompass all possible types of edits. However, by efficient solutions for insertion, replacement, deletion, translation and rotation, \methodname lays a foundation for developments in training-free 3D editing techniques. 

\section{Related work}
\noindent\textbf{Text-driven Editing in 2D}
Text-driven editing methods gained attention due to editing objects/scenes based on text prompts, thereby reducing manual labor. Recent advancements in diffusion models have notably improved the photo-realism and diversity of generated content, prompting researchers to explore their applications. InstructPix2Pix \cite{brooks2023instructpix2pix} exploits cross-attention layers mentioned in Prompt-to-Prompt \cite{hertz2022prompt}, which trains models using paired images of before/after edits. In contrast, MasaCtrl \cite{cao_2023_masactrl} introduces mutual self-attention in diffusion models. Imagic \cite{kawar2023imagic} and ObjectStitch \cite{Song_2023_CVPR} make complex semantic edits such as posture or compositional editing of multiple objects. GLIDE \cite{Nichol2021GLIDETP} edits images in a photorealistic manner using classifier-free guidance in masked portions. DiffEdit \cite{couairon2022diffedit} generates implicit masks from text instruction.

\noindent\textbf{3D Scene Editing:}
 Despite advancements in text-driven editing in 2D scenes, few methods have extended this to 3D environments using NeRF-based rendering due to limited training data. They can be divided into three categories. $i)$ Prior guided, $ii)$ Geometry/Texture editing and $iii)$ Text guided insertion.

\noindent \textbf{Prior Guided}: Some scene editing techniques necessitate human supervision for precise region identification. SKED \cite{mikaeili2023sked} utilizes human sketches from multiple viewpoints to localize edit regions. Set-the-Scene \cite{CohenBar2023SettheSceneGT} adopts a proxy scene layout to perform local and global iterative edits, ensuring scene coherence. InseRF \cite{shahbazi2024inserf} inserts objects in the scene using a bounding box. While effective, these methods require human intervention, limiting their scalability and automation potential.

\noindent  \textbf{Geometry/Texture Editing}: Contrary to earlier methods, few works try to edit the scene without any prior guidance. DreamEditor \cite{zhuang2023dreameditor} represents scenes as mesh-based neural fields and makes edits using text-to-image diffusion models in the edit region identified by the text encoder. RePaint-NeRF \cite{RePaint-NeRF} semantically selects target objects and leverages a pre-trained diffusion model to guide NeRF models in editing 3D objects. ViCA-NeRF \cite{dong2023vica}, Instruct 3D-to-3D \cite{kamata2023instruct} and SINE \cite{bao2023sine} make edits to the scene through multi-view consistent 3D texture editing using \cite{brooks2023instructpix2pix}. These methods are capable of changing the geometry and texture of existing objects but cannot insert new object into the scene.

\noindent  \textbf{Text-guided Object Insertion}: The above methods are restricted to geometry/textural editing but cannot insert/replace things in scene. Instruct-NeRF2NeRF \cite{instructnerf2023} addresses this by iterative editing a set of images to perform a single edit. Vox-E \cite{sella2023vox} learns a grid-based volumetric representation of images captured from 3D objects for making edits in 3D space. FusedRF \cite{goel2023fusedrf} performs objects and scene compositing using single RF through distillation. Control-NeRF \cite{Lazova_2023_WACV} edits and manipulates scene by inserting objects across different scenes or multiplying objects within same scene. For every edit, it needs re-training, which takes ample time, making them unfit for real-time edits.

There are quite a few fast variants of NeRF that primarily are confined to scene reconstruction but not to scene edits. Also, they fail to process complex prompts that involve open vocabulary because they do not have semantic and global information about scene and depend on image edits. NeRF models face difficulty in converging the diverse edits from image to image based on complexity of natural prompt. Moreover, the main goal of this work is to present a training-free alternative for scene editing. 

\section{Method}
\label{sec:method}
This section presents our proposed text-guided 3D object insertion and replacement method. \methodname takes a 3D scene and a text prompt as inputs specifying which object should be inserted/replaced in a scene. Input text is processed for task categorization and entity extraction. Entity extraction carried out on text helps in object generation and finding the best suitable location in the scene.  We apply transformations on the generated object for placement. As a result, our method produces a union of 3D scene mesh and a 3D object placed in the scene based on the text prompt.

\begin{figure}[t]
  \centering
   \includegraphics[width=\linewidth, clip, trim = 0cm 11cm 0cm 1cm]{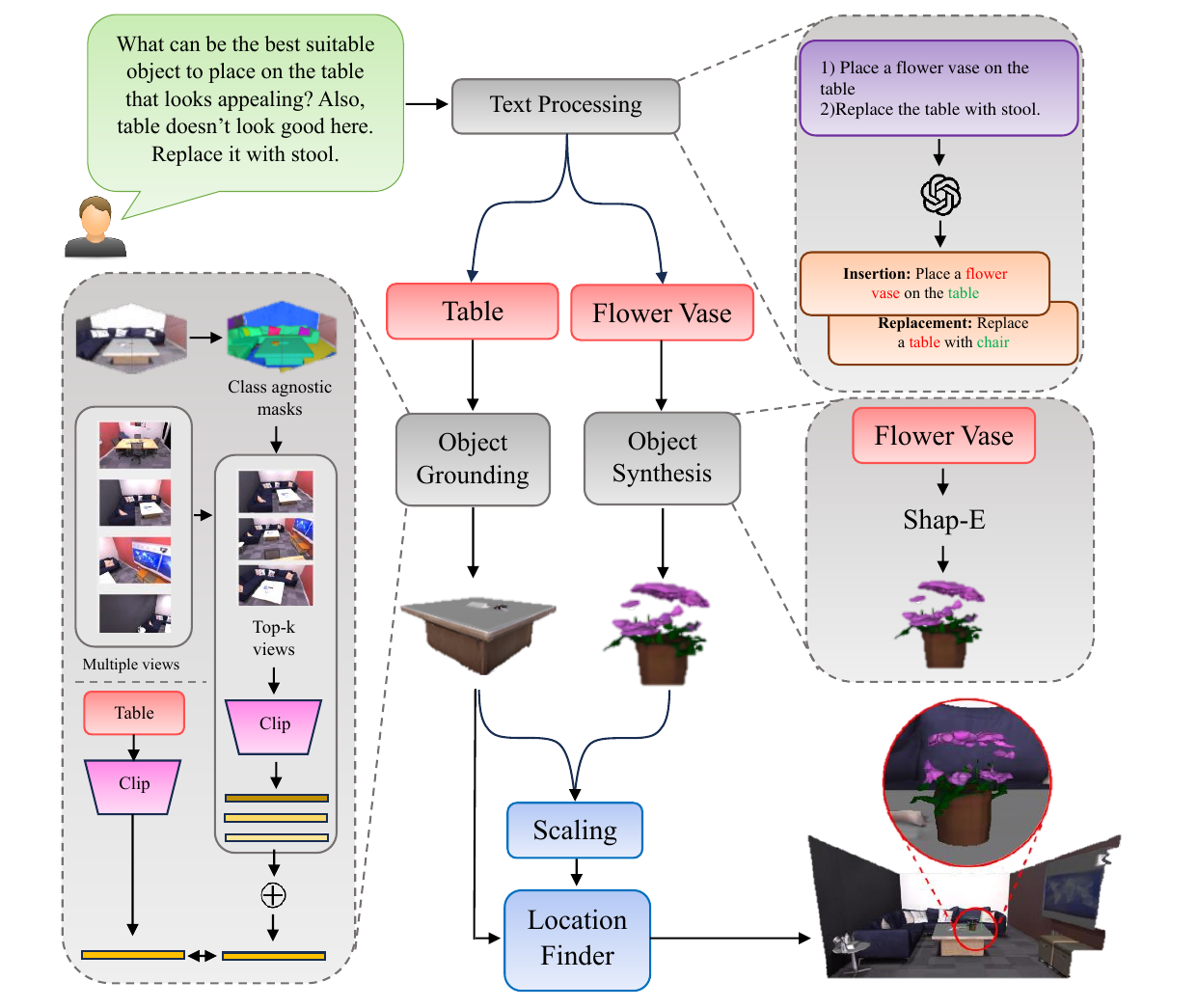}

   \caption{\textbf{\methodname}: Object insertion in a 3D scene. Given a text prompt, LLM classifies the task and extracts primary and grounding entities. Object synthesis for primary object is done by Shap-E. OpenMask3D does object grounding. Scaling of primary object is performed. Location finder computes an optimal location to place primary object on grounding object. Scaling and location finder (in blue) are not pre-trained models and run on the fly.}
   \label{fig:methodfig}
   \vspace{-0.5cm}
\end{figure}

The overview of our proposed method for object insertion is shown in Fig.~\ref{fig:methodfig}. The method includes six major steps: $i)$ Task classification and entity extraction from a given prompt using LLM, explained in Section \ref{entityextraction}, $ii)$ Text-conditioned 3D mesh synthesis of a primary object (Section \ref{objectgen}), $iii)$ Extracting the grounding object from the 3D scene 
(Section \ref{grounding}), $iv)$ Scaling the primary object with respect to the grounding object (Section \ref{scaling}), $v)$ Identifying the optimal location where the object has to be placed
in Section \ref{findingloc}, and $vi)$ Refine the placement of primary object at the identified location in Section \ref{refinement}. Similarly, last section (Section \ref{replacement}) explains the process of replacing objects.

\subsection{Preliminaries}
\label{preliminaries}
\noindent\textbf{Diffusion Models} 
\label{diffusionmodels}
Diffusion models \cite{NEURIPS2020_4c5bcfec} are the probabilistic models that gradually learn to generate data similar to the training data. They destroy training data slowly by adding noise during forward pass. In the reverse pass, they predict and remove the noise added between two consecutive steps during forward pass to generate the data. Once the model learns the denoising step, it can generate data from Gaussian noise in such a way that the generated output resembles data from the distribution on which it was trained.

\noindent\textbf{Latent Diffusion}
Latent Diffusion \cite{Rombach_2022_CVPR} is a two-step process where initially, the encoder and decoder are trained to produce latent codes from data $Z = E(x)$ and reconstruct data from these latent codes $\Tilde{x} = D(Z)$. Next, a diffusion model is trained to denoise latent representations for generating data, resulting in low computational costs with latent codes capturing the important details of data.

\noindent\textbf{Signed Distance Function (SDF)}
Signed Distance Function (SDF) is a mathematical function that represents the shape of a 3D mesh. It maps the coordinates of a point to a scalar value $f(x) = D$
which is the orthogonal distance from the point $x$ to the nearest point on the surface of the mesh. The sign of $D$ determines the interiority of the point with respect to the surface boundary, negative sign for inside points, positive for points outside the mesh and 0 for points on the surface. 
Algorithms like Marching cubes \cite{DBLP:conf/siggraph/LorensenC87} can be employed to construct the surface from SDF.

\subsection{Task Classification and Entity Extraction}
\label{entityextraction}

This section explains ``What'' needs to be inserted in a 3D scene and ``Where''. To do this, we leverage GPT-4 \cite{achiam2023gpt}, a Large Language Model (LLM) to categorize the task into Insertion/Replacement or Deletion.  For insertions, the LLM identifies primary ("what") and grounding ("where") objects from the prompt. In replacement, it determines which object to replace and its substitute. LLMs excel at interpreting natural language prompts with open vocabularies, making them invaluable for providing accurate answers and relevant suggestions. GPT-4 can process complex paragraphs by breaking them into manageable tasks, categorizing them, and extracting necessary entities. It offers context-specific recommendations, such as suggesting appropriate dishes for a kitchen or suitable decorative items for a living room. Moreover, LLMs understand practical considerations, like plants requiring sunlight, and provide advice accordingly (see Fig. \ref{fig:results}, row 2). Examples of prompts are available in the Supplementary materials.


\subsection{Object Synthesis}
\label{objectgen}
After identifying the primary object in the text prompt as in Section \ref{entityextraction}, it is used to generate a 3D object which should be placed in a scene. We use a pre-trained model for text to 3D generation mentioned in \cite{jun2023shap}. Shap-E is a transformer-based generative model built on \cite{nichol2022point}  that produces a 3D mesh, conditioned on text or an image using latent diffusion. 
 The role of this generative model is to synthesize any object as per requirement with modernized traits. It do not restrict the primary object to set of objects that can be retrieved from repository. This helps in designing primary object with diversified features whereas in object retrieval, it is limited to predefined object categories. 

\subsection{Object Grounding}
\label{grounding}
Unlike previous methods, our method reduces the user interaction by eliminating process of creating bounding box or masks (process of explicitly mentioning the grounding object). Instead, from the given scene, we extract this object using an open vocabulary grounding mechanism. Open vocabulary object grounding provides a flexibility of localizing objects without restricting to predefined set of object classes and can be deployed for unseen objects. This process is good at compiling natural languages 
 enhancing human communication. We use a pre-trained open vocabulary 3D instance segmentation model to ground the surface object in a 3D scene. OpenMask3D\cite{takmaz2023openmask3d} 
  computes feature representation of 3D objects aggregating CLIP \cite{Radford2021LearningTV} features from multi-view images and retrieves objects having high similarity with query features. The name of the grounding object obtained in Section \ref{entityextraction} is embedded using CLIP.
 The mesh of object whose representation is similar to query embedding is retrieved.

\subsection{Scaling: Primary vs Grounding Object}
\label{scaling}
The primary object is generated in a different coordinate space, and the scene is in another space. As a result, the primary object is independent of the scene and has to be scaled accordingly to orient perfectly in the given scene. To solve this, we consider 2D latent diffusion. \cite{Rombach_2022_CVPR} generates images based on text using cross-attention layers in diffusion models. 
We generate a few images based on the text prompt using 2D diffusion, and these images are passed to \cite{liu2023grounding}. Grounding DINO is a text guided transformer-based open-set object detector that detects the objects in the input image. 
We consider a set of images containing both objects so that we get the scale of one object with respect to the other. Bounding boxes (bb) of both objects are predicted in each image and the dimensions are extracted. We consider width primarily to compute the scale as width plays a major role in determining the space needed for placing. Scale ($\mathcal{S}$) is the ratio of the width of bb of the primary object ($\mathit{W}_p$) to the width of bb of the grounding object ($\mathit{W}_g$).
\begin{equation}
\label{scale}
  \mathcal{S}  = \frac{\mathit{W}_p}{\mathit{W}_g}
\end{equation}

Out of all the computed scales from generated images, we consider the minimum value and use it for scaling the primary object in a 3D scene as the results look realistic. Our model doesn't perform practical edits if the scale is of larger value. For more details, please refer supplementary.

\subsection{Location Finder}
\label{findingloc}
To minimize user interaction, we propose a technique to obtain a location where the object should be placed without user intervention. For this, we first extract the vertex normals of the grounding object retrieved from Section \ref{grounding}. 
\begin{figure}[t]
  \centering
   \includegraphics[width=\linewidth, clip, trim = 0cm 6cm 0cm 6cm,]{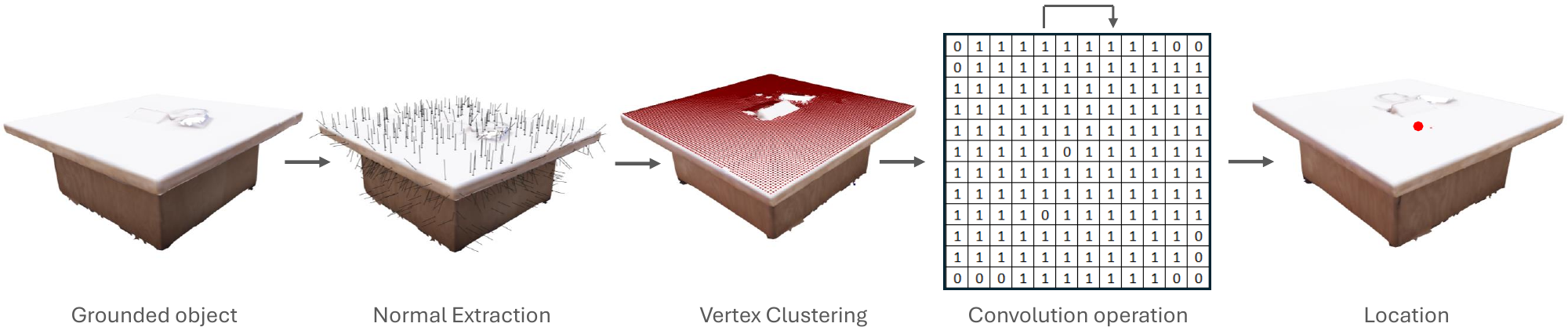}

   \caption{From grounded object on the left, vertices having normals parallel to the positive $Z$-axis are filtered. They are clustered based on density, visualized as red dots and a voxel grid is created. Convolution operation is performed on this voxel grid which finally gives the optimal location pointed with red dot.}
   \label{fig:locationfig}
\end{figure}
Please note that the our focus is only on surfaces that are parallel to the ground. We filter the vertices whose vertex normals are 
orthogonal to the ground, pointing towards the roof as shown in Fig.~\ref{fig:locationfig}. Other vertices are omitted which eliminates the regions that are unfit for placing the objects leaving cavities on the surface.
Filtered vertices are clustered using density-based clustering (DBSCAN). As a result, the vertices having nearly the same $Z$ values are clustered together. 
Width of scaled primary object is divided into $n$ voxels each of size $s$. 
Similarly, the each cluster of the grounding object containing filtered vertices is voxelized, each voxel having size $s$. This forms a grid with $M$ rows and $N$ columns on the surface of the grounding object. The value in each empty cell is filled with `0' and cells containing at least one vertex are marked with `1'. 
\begin{figure}[t]
  \centering
   \includegraphics[width=0.7\linewidth, trim = 2cm 0.5cm 2cm 1cm,height = 2.5cm]{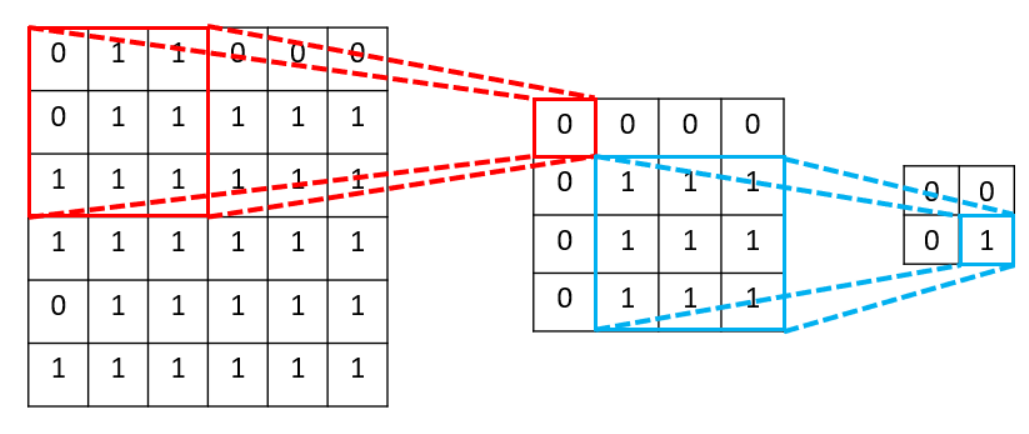}

   \caption{Level-0 voxel grid of size $6 \times 6$. `1' represents presence of at least one vertex. Filter of size $3 \times 3$ is slithered over level-0 to get level-1 with value `1' if a threshold condition is met, else `0'. Similarly, level-2 is computed from level-1 for the final output.}
   \label{fig:hierarchy}
\end{figure}

A filter of size $n \times n$ is created and slithered over the grid. Convolution operation is performed on the grid, and average operation is applied. The output of this operation is `1' if the average value exceeds the threshold; else, the output is `0' as shown in Fig.~\ref{fig:hierarchy}. Next, the filter is shifted by one cell till the entire grid is covered, and for each shift, similar operation is performed. The size of the resultant grid is $(M-n+1) \times (N-n+1)$ where $M$ and $N$ are the rows and columns of the voxel grid from previous level respectively. This process will continue until there is no `1' in the output grid or size of output grid is less than the filter size (after multiple levels). The value `1' in the output grid specifies the location containing vertices. This process enables us to find locations where there is a feasibility to place primary object and out of all the viable locations, optimal location is selected that minimizes the intersection with prior objects in scene. Finding the best placement among all possible results is a non-trivial problem, considering the text prompt might not always have detailed instructions, multiple positions of placement might be correct. Keeping this in mind, we have developed our method in such a way that it finds the best spot by avoiding mesh intersections and preferring surfaces with fewer collisions among multiple options.
As we repeat the process over multiple levels, 
the grounding surface is examined at multiple scales (obtaining global information) to locate the finest position, far from prior objects. Index of valid voxel from final grid is extracted and coordinates of prime location are calculated using 
\begin{equation}
\begin{aligned}
    (x,y) = (x_0+(x_{id} + levels - 0.5 ) * width\\
    y_0+(y_{id} + levels- 0.5 ) * width)
    \end{aligned}
\end{equation}

where $x_{0}/y_{0}$ are the minimum values along $x/y$-axis in cluster, $x_{id}/y_{id}$ are the indices of voxel along $x/y$ axis in final grid, $levels$ is number of levels in hierarchy and $width$ is voxel size. 
Neighboring vertices are considered to compute the $Z$ coordinate. Experiments with different threshold values and different filter sizes are mentioned in Supplementary.

\subsection{Refinement}
\label{refinement}
Refinement is required to determine the best orientation of the primary object to be placed on the grounding object with minimal intersection. The base centroid of the primary object is calculated. Using this centroid, the object is translated to the location identified in Section \ref{findingloc}. Now the object is rotated by 
certain angles at equal intervals and angle with least penetration percent (Eq.\ref{penetrationpercent}) is considered to rotate the object. With this, the primary object is transformed to the desired location, and both meshes are merged.

\subsection{Replacement}
\label{replacement}
Our proposed method can also replace objects in a given scene mentioned through a text prompt. Given prompt is processed by LLM as mentioned in Section \ref{entityextraction} to identify the existing object and replacing object. If a single object is mentioned in the text prompt, a similar object will replace the existing object. There is also a provision to input 3D mesh of replacing objects instead of generating new. The grounding object is extracted using \cite{takmaz2023openmask3d}, and its dimensions are recorded. This grounded object is deleted from the scene and cavities are inpainted as mentioned in \ref{deletion}. The replacing object is scaled according to the dimension of the grounding object so that it fits exactly in the place of the grounding object. After scaling, this mesh is transformed to the location of the grounded object and fused with the scene mesh. Detailed process is described in supplementary.

\section{Experiments}
\label{sec:experiments}

In this section, we discuss the details of the datasets. We mention the baseline to compare our proposed method, the metrics used and both qualitative and quantitative results in Section \ref{results}. These evaluations are mainly to find the presence of intersections between primary and other objects located on grounding objects in a scene. Moreover, we provide functionalities that can be used to refine the results in Section \ref{manoeuvre}. 

\noindent\textbf{Datasets}
\label{datasets}
We use ScanNet \cite{dai2017scannet} and Replica  \cite{replica19arxiv} datasets for our experiments. ScanNet is a real-world dataset containing $1500+$ indoor scenes with rich annotations. It has $1201$ training scenes and $312$ validation scenes and a few hidden test scenes. We consider $20$ scenes randomly from the validation set that comprises different objects such as tables, chairs, couches, etc. Replica is synthetic dataset containing $18$ high-resolution indoor scenes, each having semantic-level segmentation information. We consider $8$ scenes mentioned in \cite{takmaz2023openmask3d} for our experiments. Since, \methodname works on both synthetic and real-world datasets, we are sure that this can be applied to any type of dataset.

\begin{figure*}[t]
  \centering
   \includegraphics[width =0.9\linewidth, clip, trim = 0cm 11.8cm 0cm 0cm,height = 12.2cm]{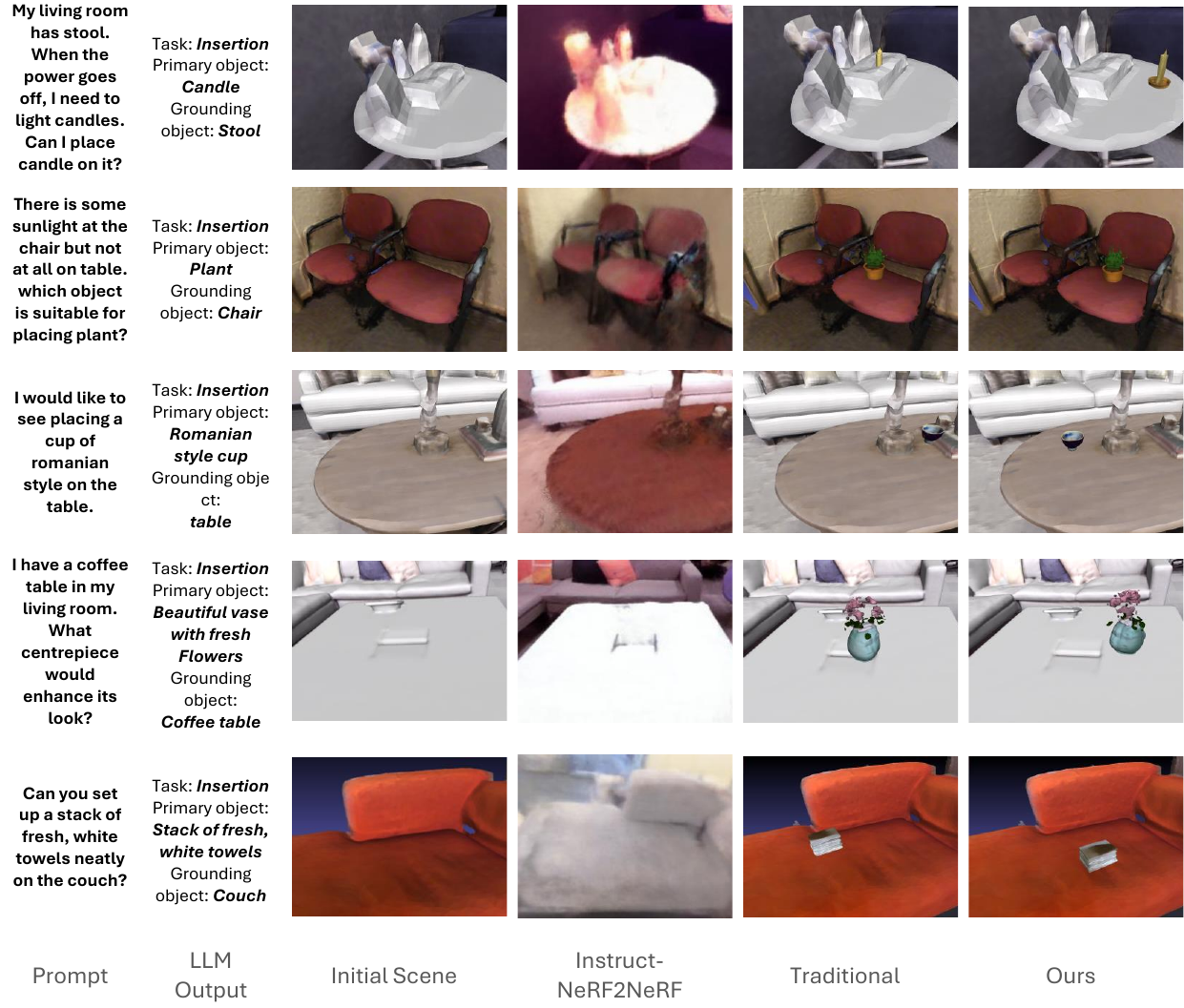}
\caption{\textbf{Insertion}. (a) The input prompt, (b) Response from LLM, (c) The input scene, (d) Output from Instruct-NeRF2NeRF, (e) Results of our traditional baseline, (f) Output of \methodname.}
\label{fig:results}
\end{figure*}

\subsection{Results}
\label{results}
We demonstrate the results of our method against the baseline methods. We consider two baselines for our main results. $i)$ Instruct-NeRF2NeRF\cite{instructnerf2023} is a recent text-guided 3D scene editing method. It uses an image-conditioned diffusion model to iteratively edit the images while optimizing the underlying scene. $ii)$ We propose another baseline similar to \methodname. It is based on pre-trained models used in our pipeline, followed by scaling of the primary object. We do not consider location finder algorithm as it is our proposed algorithm. 
Instead, we place the object by default at the center of the grounding object in baseline. Other steps are similar to \methodname as discussed in Section \ref{sec:method}.

\begin{figure}[t]
  \centering
   \includegraphics[width =\linewidth,clip, trim = 0.9cm 19.7cm 2cm 0cm]{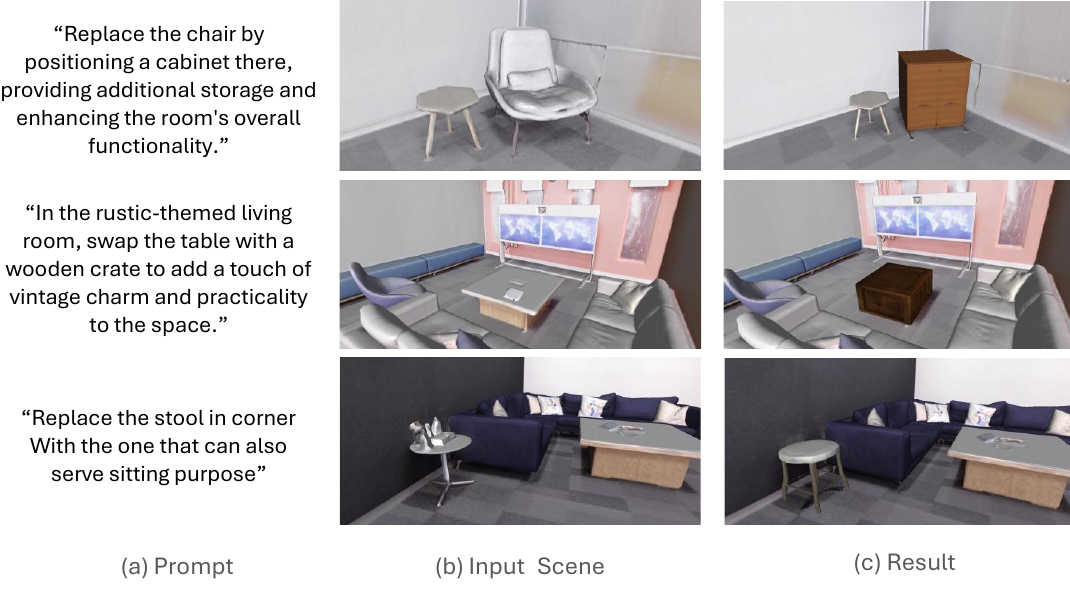}
\caption{\textbf{Replacement}. (a) The input prompt, (b) Input scene before replacement, (c) Scene after replacement using \methodname.}
\label{fig:replace}
\end{figure}
\noindent\textbf{Qualitative Results}
Our qualitative results for object insertion are shown in Figure \ref{fig:results}. We consider five samples of scenes and text prompts. We show scenes after editing by our proposed \methodname, Instruct-Nerf2Nerf and our traditional baseline where primary object is inserted on grounding object in each example. We can clearly observe that our method inserts the object at the best-suited location without any intersection with other objects, while the traditional baseline fails to avoid intersections. On the other hand, Instruct-NeRF2NeRF attempts to insert a new object in the scene, leading to global changes (rows 1,2,3) in the scene and modifying existing objects instead of inserting a new object(row 5). We can see the change in texture, color, etc. in Fig. \ref{fig:results} (d).  This is because Instruct-NeRF2NeRF performs editing in 2D images and placing things without strong semantics leads to diverse edits in images. These edits may include variations in primary object or different placing locations and Instruct-NeRF2NeRF fails at converging these edits resulting in global edits. 
 From Figure \ref{fig:results}, please note (especially in examples 1,3,4) that, compared to other methods, \methodname outperforms the insertion job by avoiding intersections with prior objects and ignoring the uneven surface, which is not suitable for placing object.
Subsequently, we show our qualitative results for object replacement in Figure \ref{fig:replace}. We can see that when a single object is mentioned in the text prompt (last row), a similar object will replace the existing object otherwise primary object replaces the grounding object. Our results indicate that our method smoothly replaces the existing object without discrepancies, intersections and distortions in the scene. For more examples, refer supplementary.

\noindent\textbf{User Study Evaluation} Next, we evaluate our results by conducting a small user study with 35 participants having research background  with age in between the range of 19-30 years. We evaluate 8 test cases following the practice of \cite{zhuang2023dreameditor,cao24dream}. 
Our questionnaire includes input prompt, edited output of the baseline and our method and questions related to these methods. The input prompt guides the user to understand the edit operation performed. The outputs of our method and baseline are ordered randomly in each questionnaire and no pattern/clue was available whatsoever. It contains questions like $a)$ ``Is primary object placed correctly on the grounding object?'', $b)$ ``Rate the look and feel of scene'', and $c)$ ``Rate the orientation of primary object''. We requested people to rate the results on a scale of $1-5$, with $5$ being excellent. We collected responses and averaged the scores for each question. We found the scores to be \textbf{4.17} and 3.02 for question $a)$ for \methodname and baseline, respectively. Similarly, $b)$ and $c)$ have scores of \textbf{3.86} and 2.78, and \textbf{4.17} and 3.16 for our method and traditional baseline (the higher the better). These results demonstrate that the quality of scenes edited by our method is better than the baseline in real-world.

\noindent\textbf{Quantitative Results}
As no ground truth information available, we considered a quantitative metric motivated by \cite{tendulkar2022flex,turpin2022grasp}. This metric calculates the percentage of primary object vertices intersecting with the scene mesh.
Penetration percent metric is the count of primary object vertices having negative signed distance divided by the total number of primary object vertices. A point will have negative signed distance if it lies inside the mesh, positive if it lies outside the mesh and 0 if it lies on the mesh. Since we insert objects inside a closed scene, intersection results in vertices penetration out of the scene leading to positive sign for penetrated vertices.
\begin{equation}
\label{penetrationpercent}
   Penetration\ percent = \frac{1}{N} \sum_{n=1}^{N} \ [dist(v_n,M) > 0]
\end{equation}

\begin{table}
\begin{center}

\begin{tabular}{lll}
\hline\noalign{\smallskip}
Refinement & \methodname & Traditional\\
\noalign{\smallskip}
\hline
\noalign{\smallskip}
No  & \textbf{4.40\%} & 8.21\%\\
Yes & \textbf{3.84\%} & 7.45\%\\
\hline
\end{tabular}
\caption{Penetration percent. Bold represents the best result.}
\label{table:penetrationpercent}
\end{center}
\end{table}

where $N$ is total number of vertices in primary object, $v_n$ is the $n^{th}$ vertex of primary object, $M$ is scene mesh and $dist(x,y)$ is a Signed Distance Function (SDF).
For this experiment, we consider $50$ grounded objects from $28$ scenes and generated $25$ primary objects, such as a coffee cup, teapot, laptop, etc. With this, the whole evaluation set contains $1250$ cases. We evaluate our proposed method against the traditional baseline on all the cases using the penetration percent metric. We observe that the average penetration percent value is reduced by nearly half from the baseline method in the proposed method as shown in Table~\ref{table:penetrationpercent}. This indicates that the percent of vertices intersecting with scene mesh is reduced by significant margin using our proposed location finder algorithm even without refinement step. After refinement (Section \ref{refinement}), the percentage is further reduced for both the methods.

\subsection{Manoeuvre}
\label{manoeuvre}
In this experiment, we provide an option for further refining the placement of object. We can translate, rotate, delete or iteratively add objects in a scene.

\begin{figure}[t]
  \centering
   \includegraphics[width = \linewidth, clip,trim = 3cm 19.8cm 3cm 2cm]{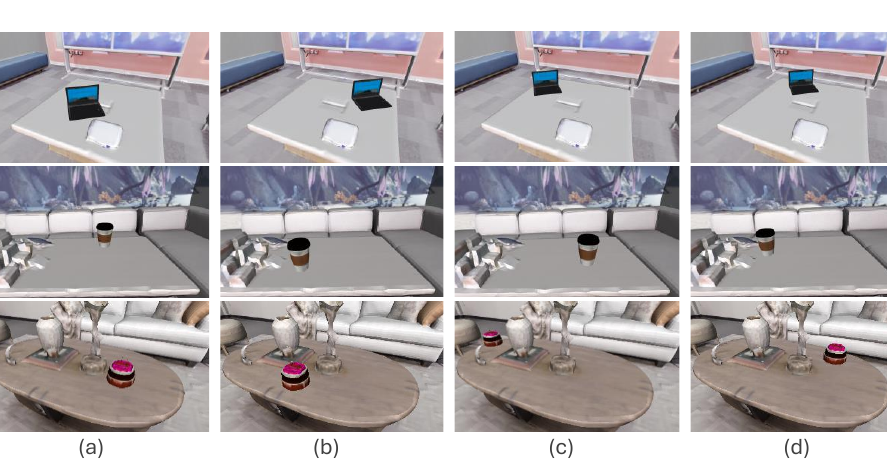}
\caption{\textbf{Translation}. (a) The initial placement of primary object,  (b), (c) and (d) The translation of the object to different points. }
\label{fig:translation}
\vspace{-0.5 cm}
\end{figure}
\noindent\textbf{Translation}
\label{translation}
If the position of the primary object has to be changed, we can select a point on the grounding object. Now the object will be translated to the point we select. If more than one point is selected, the mean of all these points is calculated, and the primary object is translated to the centroid point. Few examples are demonstrated in Fig.~\ref{fig:translation}.

\begin{figure}[!h]
\centering
\includegraphics[clip,width = \linewidth,  clip,trim = 3cm 19.8cm 3cm 2cm]{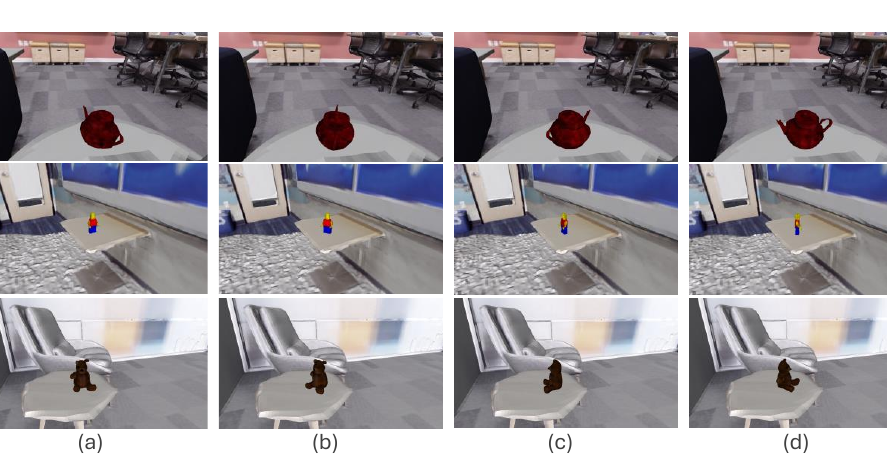}
\caption{\textbf{Rotation}. (a) The initial placement of the primary object, (b) Primary object rotated by 45 degrees in clockwise direction, (c) Object rotation by 90 degrees in clockwise direction and (d) Object rotation by 60 degrees in anticlockwise direction.}
\label{fig:rotate}
\end{figure}

\noindent\textbf{Rotation}
\label{rotation}
Similar to translation, we can rotate the placed object accordingly. We can rotate the object in a clockwise or anticlockwise direction from the top view by specifying the angle. Given the angle and direction, the primary object is rotated around the $Z$-axis at the placed location on the grounding object. Fig.~\ref{fig:rotate} illustrates a few examples.

\begin{figure}

  \centering
  \includegraphics[clip, trim = 3.2cm 19cm 6cm 1cm,width = 0.9\linewidth,height = 5.5cm]{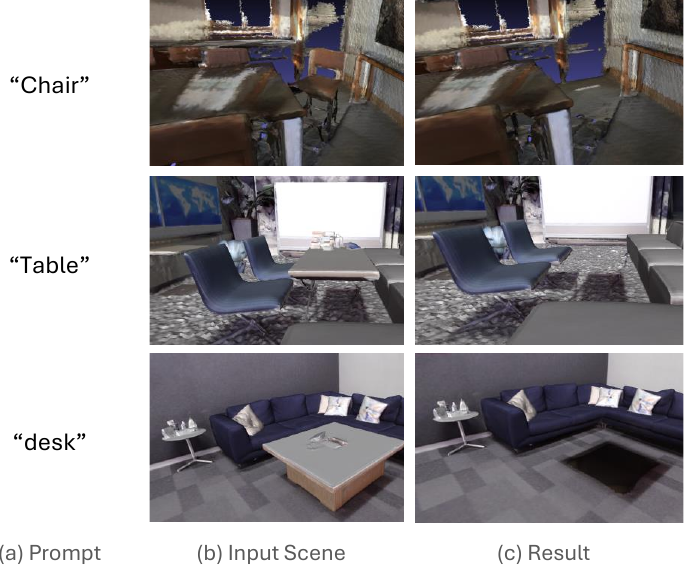}
  \caption{\textbf{Deletion}. (a) The text prompt, (b) The input scene, (c) Scene after deletion.}
  \label{fig:delete}
  \vspace{-0.4cm}
  \end{figure}
\begin{figure}
\vspace{-0.8cm}

  \centering
  \includegraphics[clip, trim = 2.5cm 7.5cm 2.5cm 0cm,width = 0.9\linewidth]{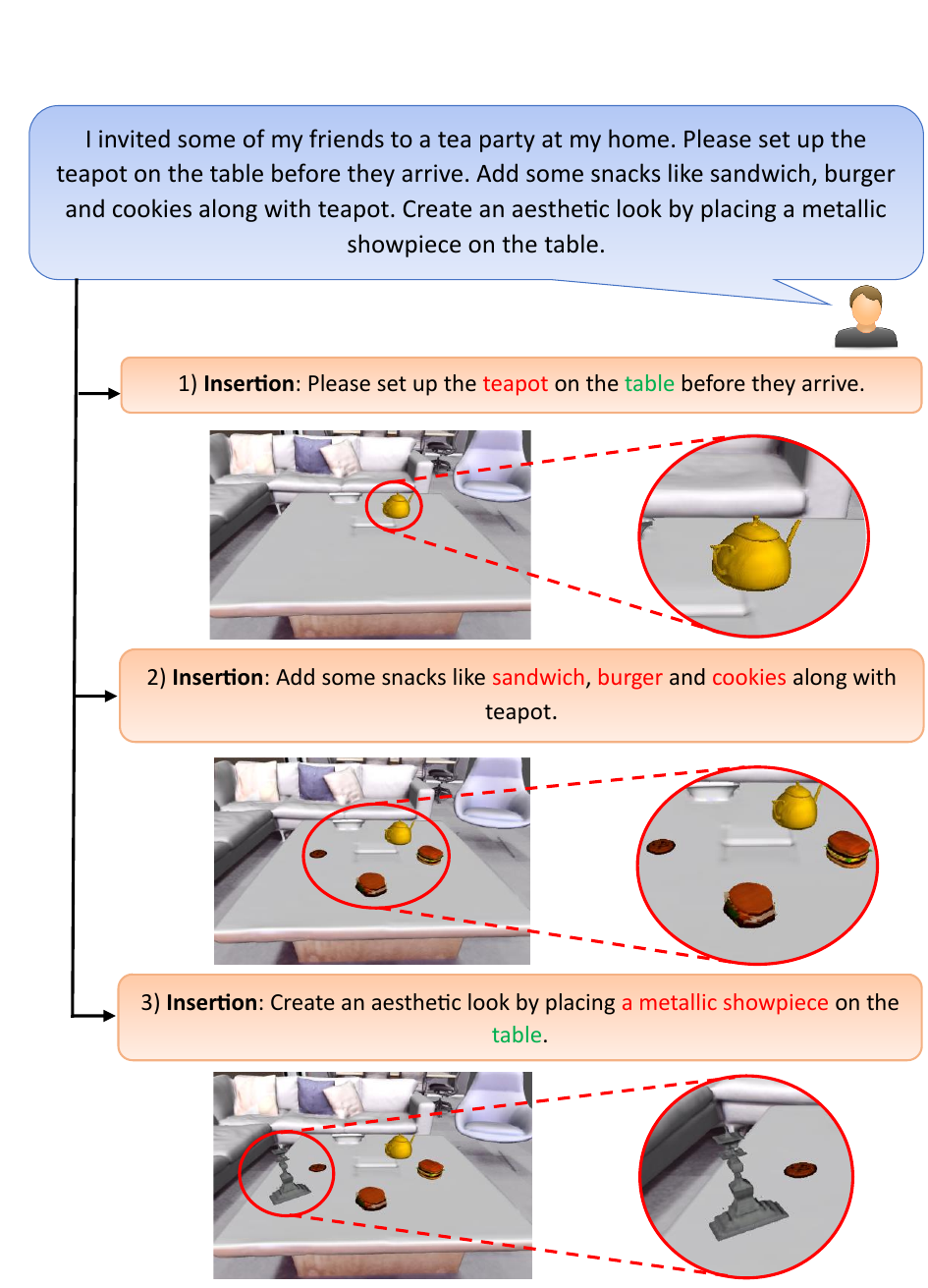}
  \caption{\textbf{Iterative Insertion}. Given a text prompt, objects are added onto grounding surface in iterative fashion.}
  \label{fig:iteradding}
\vspace{-0.3cm}
\end{figure}

\noindent\textbf{Deletion}
\label{deletion}
Given a text prompt, we can delete existing objects in a 3D scene as shown in Fig.~\ref{fig:delete}. Based on the feed, the related object is grounded using OpenMask3D \cite{takmaz2023openmask3d}, and it is deleted from the scene. \methodname deletes other objects placed on the grounding object as well. Deletion of objects results in the cavity in the scene. A new surface is created connecting boundary vertices of cavity but the faces in the new mesh are much larger leading to inappropriate shading. This issue can be tackled by breaking down the larger faces into small and each vertex feature can be computed considering nearest neighbors. Also note that appearance tailoring is out of the scope of this paper.

\noindent\textbf{Inserting Objects Iteratively}
\label{iterative adding}
\methodname is capable of adding multiple objects iteratively. When our model is provided with prompts containing multiple primary objects, it adds them in a recursive manner, each object at a time. Initially, the first object is generated and placed on the grounding object, and the next object is processed. This process continues until either all the objects are added or there is insufficient space to place an object. The model makes sure that new object does not intersect with other objects. 
 This ensures intersection-free placement of objects in an iterative fashion. Results are shown in the Fig.~\ref{fig:iteradding}


\noindent\textbf{Limitations and Future work:}
\label{limitations}
Following are some of the limitations of \methodname that present opportunities for future work. $i)$ Object placement is currently restricted to flat surfaces. Incorporating physics simulations guided by LLMs could enable placement on non-flat surfaces. $ii)$ The deletion of objects can result in illegitimate inpainting within cavities (as discussed in Section \ref{manoeuvre}). $iii)$ The system lacks the ability to place objects precisely in relation to existing objects (e.g., ``in front of"). This limitation could be addressed by incorporating scene graphs to provide spatial relationship information. $iv)$ Unrealistic object placement occurs when scaling values exceed certain thresholds. This issue can be mitigated by using appropriate threshold limits.


\section{Conclusion}
\label{conclusion}
We introduced \methodname, a text-guided approach for adding and replacing objects in 3D scenes comprising multiple objects. Given a scene along with a prompt for editing, the proposed method extracts the entities and synthesizes the object to be inserted (or replaced), followed by identification of the object to anchor it to. The optimal position on the grounding object is automatically determined and the synthesized object is placed. To the best of our knowledge, our method is the first text-guided training-free approach to edit 3D scenes with multiple objects. We show how leveraging mesh-level representation for simple edits can be a potential alternative to training-based and time-consuming NeRF based approaches. We showcase the effectiveness of our approach through qualitative and quantitative comparisons with relevant baselines and existing approaches.

\noindent\textbf{Acknowledgments:} This work is supported by Fujitsu Research of India Private Limited. We also thank the participants for participating in the user study evaluation.


{\small
\bibliographystyle{ieee_fullname}
\bibliography{egbib}
}

\end{document}